\documentclass[10pt,twocolumn,letterpaper]{article}

\usepackage[pagenumbers]{cvpr} 

\definecolor{cvprblue}{rgb}{0.21,0.49,0.74}
\usepackage[pagebackref,breaklinks,colorlinks,allcolors=cvprblue]{hyperref}
\usepackage{multirow}
\usepackage{pifont}
\usepackage{xparse}
\usepackage{microtype}
\usepackage{animate}

\newcommand\blfootnote[1]{ 
  \begingroup
  \renewcommand\thefootnote{}\footnote{#1} %
  \addtocounter{footnote}{-1} %
  \endgroup
}

\def\modelname{MetaShadow}
\def\stageonename{Shadow Analyzer}
\def\stagetwoname{Shadow Synthesizer}

\title{\vspace{-8mm} MetaShadow: Object-Centered Shadow Detection, Removal, and Synthesis \vspace{-5mm}}

\author{Tianyu Wang$^{1,2,}$\thanks{}, Jianming Zhang$^{1}$, Haitian Zheng$^{1}$, Zhihong Ding$^{1}$, Scott Cohen$^{1}$,\\ Zhe Lin$^{1}$, Wei Xiong$^{1}$, Chi-Wing Fu$^{2}$, Luis Figueroa$^{1,\dagger}$, Soo Ye Kim$^{1,}$\thanks{}\\
$^1$ Adobe Research \\
$^2$ The Chinese University of Hong Kong \vspace{-5mm}
}

\begin{document}
\tracingall
\twocolumn[{
\renewcommand\twocolumn[1][]{#1}
\maketitle
\begin{center}
    \vspace{-20pt}
    \includegraphics[width=0.75\linewidth]{fig/Teaser.pdf}  
    \vspace{-5pt}
    \captionsetup{type=figure}
    \caption{%
        \textbf{{\modelname}} is a versatile three-in-one framework designed for shadow-related tasks, enabling shadow manipulation in various object-centered image editing operations such as: 
    [I] Object Relocation: Our model can detect and remove the shadow of an existing object, then synthesize the shadow in the new location consistent with the original shadow.
    [II] Remove an object and its shadow: (1) Based on the mask of the unwanted object, our model can directly remove its shadow (2). After removing the object (3),
    we can eliminate any remaining shadows for a cleaner background (4) if we do not specify which shadow to remove.
    [III] Insert an object and synthesize its shadow: When inserting the person in (b) to another image (a) with similar lighting, our model can generate a realistic shadow, enhancing the final compositing quality.
}
    \label{fig:teaser}
\end{center}
}]

\blfootnote{* Work done during an internship at Adobe.}
\blfootnote{$\dagger$ Co-corresponding authors.}

\vspace{-5mm}
\begin{abstract}
{Shadows are often under-considered or even ignored in image editing applications, limiting the realism of the edited results. In this paper, we introduce  MetaShadow, a three-in-one versatile framework that enables detection, removal, and controllable synthesis of shadows in natural images in an object-{centered} fashion.  MetaShadow combines the strengths of two cooperative components: Shadow Analyzer, for object-centered shadow detection and removal, and Shadow Synthesizer, for reference-based controllable shadow synthesis. Notably, we optimize the learning of the intermediate features from Shadow Analyzer to guide Shadow Synthesizer to generate more realistic shadows that blend seamlessly with the scene. Extensive evaluations on multiple shadow benchmark datasets show significant improvements of  MetaShadow over the existing state-of-the-art methods on object-centered shadow detection, removal, and synthesis.  MetaShadow excels in image-editing tasks such as object removal, relocation, and insertion, pushing the boundaries of object-centered image editing.}
\end{abstract}
    
\section{Introduction}
\label{sec:intro}

Shadows play a vital role in revealing 
the realism of an image, providing strong cues on the perception of the 3D 
space and the spatial relations between objects in the
environment. However, when handling object-related image-editing tasks, such as unwanted object removal, object relocation, and object insertion,
existing applications (\eg, Google Magic Eraser~\cite{google2023magic}) in this field often simply 
neglect manipulating the shadows, greatly diminishing the overall visual coherence and realism of the edited images.

To effectively support image editing with shadows, as shown in Fig.~\ref{fig:teaser},
we need to collectively deal with three tasks: shadow detection, shadow removal, and shadow synthesis in an \textit{object-centered fashion}.
Our \textit{object-centered} approach indicates that we primarily focus on instance-level object manipulation with applications to image editing workflows. Thus, when we edit objects in an image, each object should be associated with the shadow cast by itself onto the environment such that its associated shadow can be naturally manipulated together.

As shown in Tab.~\ref{tab:comparison}, most existing works, however, treat the three tasks separately,
with several methods overlooking
the need for object-centric formulations to assist image-editing workflows.
General shadow detection~\cite{zhu2021mitigating, zhu2022single,Yang_2023_ICCV} 
predicts a single binary mask for all shadows, 
while Wang~\etal~\cite{Wang_2020_CVPR,Wang_2022_TPAMI} 
detects 
object-shadow pairs, akin to instance segmentation.
A line of shadow removal works~\cite{Le2019Shadow,wan2022style,zhu2022efficient} require binary shadow masks as input, relying on off-the-shelf shadow detectors or user-given masks, both of which can be error-prone, to remove the shadows.
Further, more advanced shadow removal methods~\cite{wang2018stacked,cun2020towards,liu2023decoupled} simultaneously detect and remove {\em all\/} image shadows, 
but do not support object-centered editing.
In contrast, shadow synthesis aims to generate shadows for objects inserted into a new scene. One line of work~\cite{Zhang_2021_CVPR,sheng2022controllable,Sheng_2023_CVPR} require additional estimated lighting or geometric parameters to produce convincing results. Another line of research~\cite{hong2022shadow,Liu2024SG} utilizes another object or its shadow as a reference for synthesizing shadows. However, the absence of an effective shadow knowledge extractor prevents these methods from producing accurate shadow shapes.
Recently, ObjectDrop~\cite{winter2024objectdrop} introduced a bootstrap supervision strategy to generate shadow synthesis data by training an object/shadow removal model.   

Intuitively, all shadow tasks are inherently related, and 
should benefit from shared knowledge.
For example, perfectly removing a shadow implies that one can derive an accurate shadow mask (shadow detection) from the image.
Furthermore, it indicates that crucial properties like softness and intensity of the shadow have been learned. Meanwhile, knowing how to predict a binary shadow mask of an object would indicate implicit knowledge of where to synthesize the object shadow if it were not present.
By design, handling each task separately limits each specialized model from benefitting from the shared knowledge
in the shadow formation cycle and hinders them from achieving higher-quality results.
Although we may apply multiple existing methods sequentially in the image editing pipeline, e.g., object relocation, this may lead to inconsistent outcomes.
This is because existing methods for different shadow tasks 
do not share common shadow knowledge, leading to suboptimal visual quality due to discrepancies in shadow shape, color, and intensity.

\begin{table}[tp]
    \centering
    \renewcommand\tabcolsep{10.0pt}
    \resizebox{\columnwidth}{!}{    \begin{tabular}{@{}lccc|cc@{}}
\toprule
\multirow{2}{*}{Method}       & \multicolumn{3}{c}{Task}          & \multicolumn{2}{c}{Condition}     \\ \cmidrule(l){2-6} 
                              & Detection & Removal   & Synthesis & Object-Centered & Reference-Based \\ \midrule
SILT~\cite{Yang_2023_ICCV}    & \ding{51} & \ding{55} & \ding{55} & \ding{55}       & \ding{55}       \\
SSISv2~\cite{Wang_2022_TPAMI} & \ding{51} & \ding{55} & \ding{55} & \ding{51}       & \ding{55}       \\
DHAN~\cite{cun2020towards}    & \ding{51} & \ding{51} & \ding{55} & \ding{55}       & \ding{55}       \\
BMNet~\cite{zhu2022bijective} & \ding{55} & \ding{51} & \ding{51} & \ding{55}       & \ding{55}       \\
ShadowDiffusion~\cite{guo2023shadowdiffusion} & \ding{55} & \ding{51} & \ding{55} & \ding{55} & \ding{55} \\
Zhang\etal~\cite{Zhang_2021_CVPR}              & \ding{55} & \ding{51} & \ding{55} & \ding{51} & \ding{55} \\ 
PixHt-Lab~\cite{Sheng_2023_CVPR}              & \ding{55} & \ding{55} & \ding{51} & \ding{51} & \ding{55} \\ 
SGRNet~\cite{hong2022shadow}  & \ding{55} & \ding{55} & \ding{51} & \ding{51}       & \ding{51}       \\ 
SGDiffusion~\cite{Liu2024SG}  & \ding{55} & \ding{55} & \ding{51} & \ding{51}       & \ding{51}       \\
ObjectDrop~\cite{winter2024objectdrop}     & \ding{55} & \ding{51} & \ding{51} & \ding{51}       & \ding{51}       \\
\midrule
\textbf{{\modelname} (Ours)}    & \ding{51} & \ding{51} & \ding{51} & \ding{51}       & \ding{51}       \\ \bottomrule
\end{tabular}}
        \caption{
    SOTA shadow-related methods and their supported task(s). Existing works handle up to two shadow-related tasks at once, with select models supporting an object-centered approach. Only one model uses other object-shadow pairs as a reference for shadow synthesis, avoiding the need for additional parameters that other models require.
        {\modelname} is a three-in-one framework that handles object-centered shadow detection, removal, and reference-based synthesis.}
    \label{tab:comparison}
    \vspace{-5mm}
\end{table}

In this work, we propose a three-in-one framework named \textbf{\modelname}, consisting of two synergistic components that enable object-centered shadow detection, removal, and synthesis simultaneously. Motivated by the limitation of specialized shadow-related models, we propose a novel training mechanism that successfully shares the shadow information across its task-specific components to achieve superior results.
To the best of our knowledge, {\modelname} is the first framework that can jointly handle all three shadow tasks in an object-centered fashion, benefitting from the shared knowledge to achieve SOTA results. 

We evaluate {\modelname} on three real-world tasks and four benchmarks. 
Our {\modelname} improves the mIoU from 55.8 to 71.0 for shadow mask detection, improves the bbox PSNR by 8.7dB for shadow removal, and reduces the local RMSE from 51.73 to 36.54 for shadow synthesis.

To summarize, the main contributions of this work are:
\begin{itemize}

\vspace*{-0.1mm}
\item
\textbf{Three-in-one Framework}: 
{\modelname} adopts 
a novel object-centered GAN with reference-based diffusion to address the challenges of shadow understanding and manipulation to achieve object-centered image editing.

\item
\textbf{Shadow Knowledge Transfer}: Our approach is the first to 
utilize shadow-rich intermediate features from a GAN to guide the diffusion, 
significantly enhancing the visual quality and controllability of shadow synthesis. 

\item 
\textbf{Task-Specific Datasets for Shadow Editing}: We build a synthetic training set (MOS dataset) for shadow detection, removal, and synthesis, along with two real-world test sets, Moving DESOBA and Video DESOBA, for thorough qualitative and quantitative evaluation in target scenarios.

\item 
\textbf{SOTA Performance on Three Tasks}: 
Extensive experiments on the benchmarks show that our method outperforms the baselines for object-centered shadow detection, removal, and synthesis.

\end{itemize}

\section{Related works}
\label{sec:rw}

\subsection{Shadow Detection}

{\bf General Shadow Detection.} Existing works in this category
aim to simultaneously detect {\em all\/ shadow pixels in an image,} producing a single \emph{general} shadow mask. The advent of deep learning revolutionized the task with CNNs~\cite{khan2014automatic,shen2015shadow, vicente2016large, nguyen2017shadow, wang2018stacked,le2018a+d, zhu2018bidirectional,Hu_2018_CVPR, zheng2019distraction,chen2020multi, zhu2021mitigating, hu2021revisiting, zhu2022single, fang2021robust, zhou2022shadow, wu2022light, jie2022rmlanet}, enabling automatic feature extraction that largely enhanced classification
performance. 
This area remains active, as recent work~\cite{Yang_2023_ICCV} increased the detection performance in shadow datasets like SBU~\cite{vicente2016large} by refining noisy labels through iterative label tuning.

{\bf Instance Shadow Detection.} Wang~\etal~\cite{Wang_2020_CVPR,Wang_2021_CVPR,Wang_2022_TPAMI} introduced a new task to detect shadows at an instance level. These models aim to detect object-shadow pairs in a scene, leveraging complex modules to form precise associations between objects and their corresponding shadows. 
This line of work can support object-centered image editing but need some additional post-processing.
\subsection{Shadow Removal} 
{\bf Explicit Shadow Removal.}
A large portion of shadow removal methods~\cite{Le2019Shadow,liu2021shadow,wan2022style,zhu2022efficient,zhu2022bijective,Sen_2023_WACV,guo2023shadowformer} require a \textit{shadow mask} as input to explicitly guide the model to remove the shadow pixels indicated by the mask. The performance of these models highly depends on the input, as an imperfect shadow mask may negatively affect the removal quality.
Recently, Guo~\etal~\cite{guo2023shadowdiffusion} proposed a diffusion-based method with an embedded shadow mask refinement branch that refines the input to improve the shadow removal quality.

\vspace{-3mm}
\paragraph{Blind Shadow Removal.}
More complex shadow removal works attempt to simultaneously detect and remove the shadows in the scene. This formulation allows for the blind removal of shadows and avoids the dependency on a shadow mask input. 
Early works by Qu~\etal~\cite{qu2017deshadownet} and Wang~\etal~\cite{wang2018stacked} introduced end-to-end network architectures combining the two tasks.
Successive works~\cite{ding2019argan,hu2019direction,cun2020towards,chen2021canet,jin2021dc,Yucel_2023_WACV,liu2023decoupled} propose new architectures to boost 
the removal quality. While extensive works~\cite{qu2017deshadownet,wang2018stacked,Le2019Shadow,liu2021shadow,wan2022style,zhu2022efficient,zhu2022bijective,Sen_2023_WACV,guo2023shadowformer,ding2019argan,hu2019direction,cun2020towards,chen2021canet,jin2021dc,Yucel_2023_WACV,liu2023decoupled,guo2023shadowdiffusion} focus on removing general shadows in a scene, Zhang~\etal~\cite{Zhang_2021_CVPR} proposed a method to remove an object and its associated shadow,  requiring lighting, geometry, and rendering parameters as additional input to achieve realistic results.

\subsection{Shadow Synthesis}
Evidently, shadow detection and removal are extensively represented in literature, however, shadow synthesis is a relatively underexplored task in the natural image domain.
Some methods~\cite{zhang2019shadowgan,liu2020arshadowgan} have been proposed to synthesize shadows for objects in virtual environments for AR applications. 
Recently, Sheng~\etal~\cite{sheng2021ssn,sheng2022controllable,Sheng_2023_CVPR} focuses on user-driven soft shadow and reflection synthesis, considering environmental variables like light and camera position. Further,~\cite{hong2022shadow} introduced SGRNet and its associated DESOBA dataset, the first work to demonstrate object-centered shadow generation using object-shadow pair references without requiring explicit light parameters.
SGDiffusion~\cite{Liu2024SG}, expands on the earlier DESOBA dataset 
to encompass 21.5K images and adopts ControlNet~\cite{Zhang_2023_ICCV} with a shadow intensity module to improve object-centered shadow synthesis quality. However, its shadow generation quality largely depends on other objects in the scene, which requires off-the-shelf shadow detection models to retrieve the additional input data.

\subsection{Joint Frameworks}
Some works~\cite{qu2017deshadownet,wang2018stacked,ding2019argan,hu2019direction,cun2020towards,chen2021canet,jin2021dc,liu2023decoupled} combine shadow detection and removal, while others~\cite{ding2019argan,hu2019mask,cun2020towards,liu2021shadow,zhu2022bijective} 
employ shadow generation for data augmentation 
to improve the shadow removal quality. Furthermore, recent works leveraging diffusion models~\cite{objectstitch, winter2024objectdrop} for holistic approaches such as object removal or insertion handle shadows in an implicit manner, forfeiting any controllability on the objects' effects on the scene, which is crucial in image-editing applications.

Yet, none of the existing works handle three object-centered shadow tasks jointly in a knowledge-sharing and mutually beneficial manner, resulting in a limited performance. In contrast,
by jointly performing object-centered shadow detection, removal, and synthesis, {\modelname} can 
greatly boost performance on shadow detection and removal,
while our shadow knowledge transfer mechanism leads to more realistic and consistent shadow synthesis.

\begin{figure}[t]
	\centering
	\begin{minipage}[t]{0.96 \linewidth}
		\centering
  \includegraphics[width=0.95\linewidth]{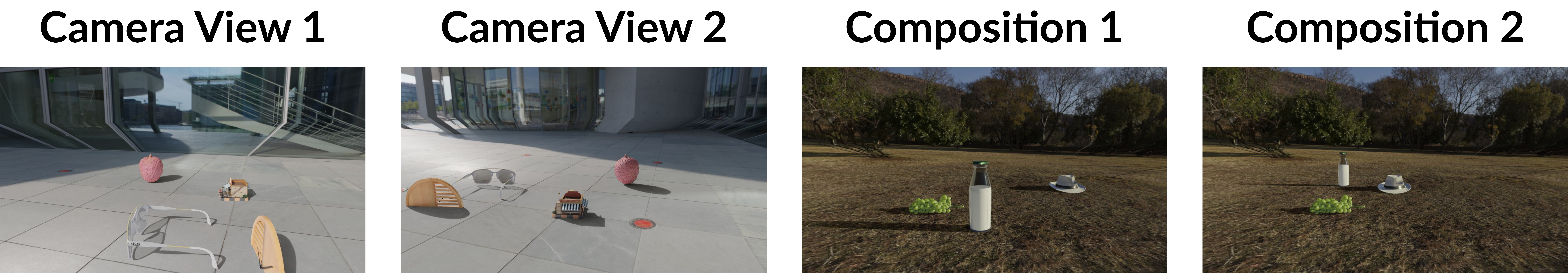}
		\vspace*{1mm}
		\centerline{\footnotesize (a) Moving With Shadow (MOS) Dataset}
	\end{minipage}
	\begin{minipage}[t]{0.46\textwidth}
		\centering
  \includegraphics[width=0.95\textwidth]{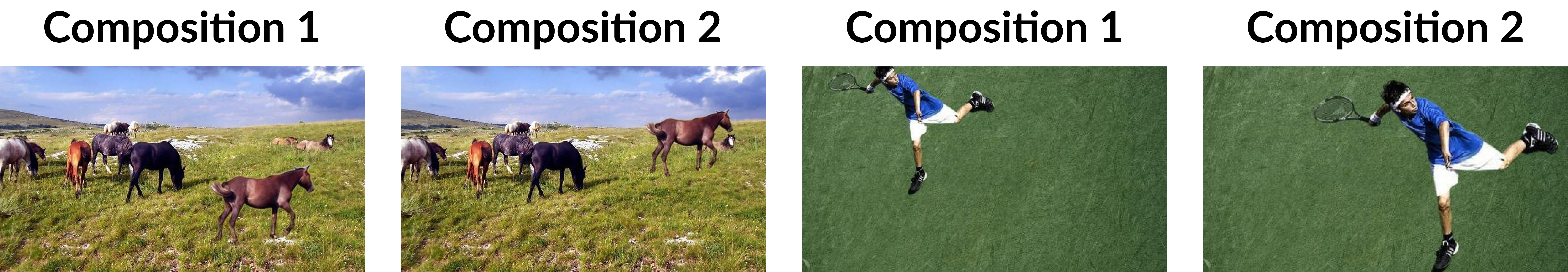}
	
		\centerline{\footnotesize (b) Moving DESOBA Dataset}
	\end{minipage}
	\begin{minipage}[t]{0.46 \textwidth}
	\centering
 \includegraphics[width=0.95\textwidth]{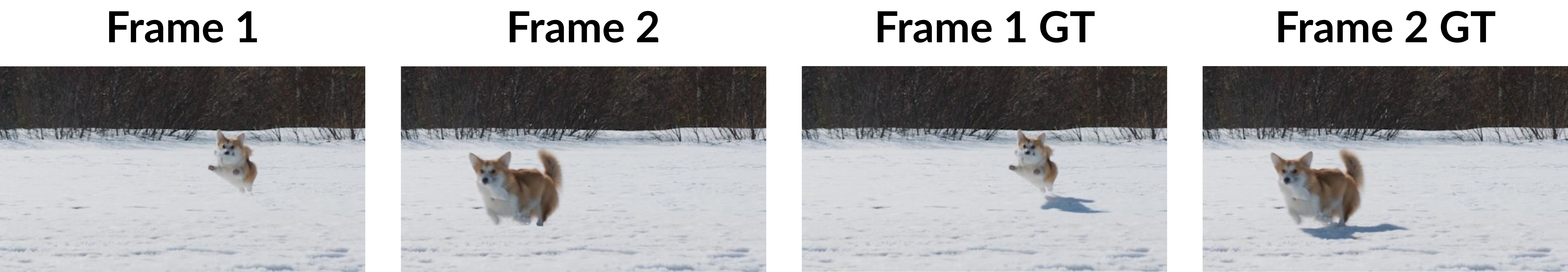}

		\centerline{\footnotesize (c) Video DESOBA Dataset }
	\end{minipage}
	
	\vspace*{-2mm}
	\caption{ We construct one training set,~\ie, MOS Dataset, and two real-world evaluation sets,~\ie, Moving DESOBA Dataset (without ground truths) and Video DESOBA Dataset (with ground truths), to train and evaluate the effectiveness of {\modelname}.}
	\label{fig:datasets}
	\vspace*{-2mm}
\end{figure}

\begin{figure*}[tp]
    \centering
	\includegraphics[width=0.9\linewidth]{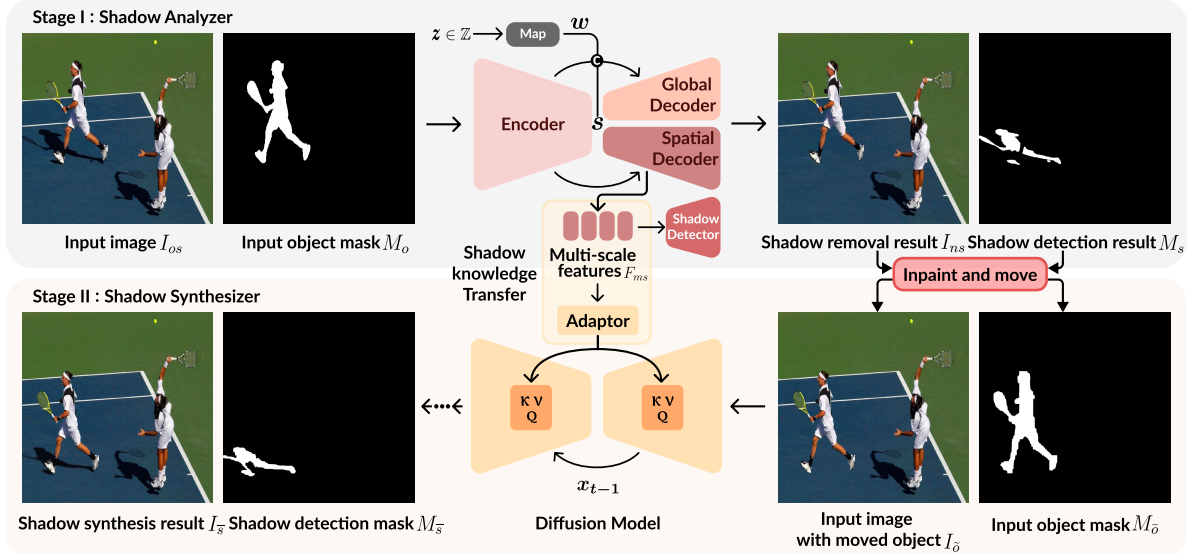}
  \vspace{-2mm}
	\caption{ The schematic illustration of our MetaShadow framework. In Stage I, the Shadow Analyzer takes the input image with object mask (left player) to perform object-centered shadow detection and removal. After that, the selected player, together with the detected shadow region, will be moved to a new location. Our Stage II then takes these as input and synthesizes {a} shadow for this object. To achieve realistic shadow synthesis, we transfer the shadow knowledge extracted from the Shadow Analyzer to Shadow Synthesizer as reference. Note that $\boldmath{s}$ represents the global style code, $\boldmath{w}$ denotes the intermediate latent space, and "K Q V" stand for key, query, and value in UNet's cross attention layer. 
}
 \vspace{-3mm}
	\label{fig:pipeline}
\end{figure*}

\section{Datasets}
\label{sec:datasets}

Contemporary shadow-related datasets~\cite{vicente2016large,wang2020instance,hu2021revisiting,Le2019Shadow,qu2017deshadownet,hong2022shadow,liu2020arshadowgan}
were built for
specific shadow tasks. 
Paired data 
preparation
for 
shadow removal and shadow synthesis are notably challenging and expensive to collect, leading to the scarcity of real-world datasets.
Additionally, only the DESOBA~\cite{hong2022shadow} and Shadow AR~\cite{liu2020arshadowgan} datasets support object-centered shadow detection, removal, and synthesis. 
Yet, DESOBA provides only 840 images for training and Shadow-AR provides only 13 3D models for rendering. Their limited scale can severely  limit a model's generalizability. Moreover, neither dataset provides samples for object relocation, which is a highly demanded image editing.

To address the limitations of existing datasets for model training, we introduce the \textbf{Moving Object with Shadow (MOS) Dataset}, synthesized using the Blender Cycles rendering engine~\cite{blender2023blender}, shown in Fig.~\ref{fig:datasets} (a). The dataset consists of 200 scenes, each with eight camera views. In addition, there are five object relocation cases for each scene, resulting in a total of 8,000 image/ground truth pairs.
Furthermore, to evaluate the applicability of {\modelname} for real-world scenes,
we also introduce two evaluation sets:
(i) \textbf{Moving DESOBA} and (ii) \textbf{Video DESOBA}. (i) For each image in the DESOBA test set~\cite{hong2022shadow}, we randomly choose an object and reposition it to
a different location; see Fig.~\ref{fig:datasets} (b) for examples.
(ii) This test set consists of twelve tripod-captured videos with static backgrounds, featuring moving objects casting shadows. Examples are shown in Fig.~\ref{fig:datasets} (c).
We will release Moving DESOBA and Video DESOBA for future evaluation; see Supplementary Materials for details.

\section{Methodology}
\label{sec:method}

As shown in Fig.~\ref{fig:pipeline}, we design two cooperative components in {\modelname}: 
(i) \textbf{{\stageonename}}, an object-centered GAN model that jointly detects and removes an object's shadow 
by taking an \textit{object mask} and an RGB image as input,
and (ii) \textbf{{\stagetwoname}}, a reference-based diffusion model that synthesizes shadows using an object mask from the {\stageonename}.

A reference object casting a shadow is often available in image editing scenarios involving natural images.
For object insertion, some shadows may already exist
in the scene. 
For object relocation, if the original object shadow is available, such shadow can be used as a reference to improve the consistency in the edited 
result.
Furthermore, the reference shadow can be a way to manipulate the generated shadow as desired, which can be useful for creative editing.

\subsection{{\stageonename}}
{Unlike existing shadow removal models that remove shadows within regions in the \textit{shadow mask}, our Shadow Analyzer needs a higher level understanding of the image, especially on the lighting and geometry of the scene,
so as to enable it to \textit{identify the object shadow} 
{and remove it}.
To this end, we base our model architecture on CM-GAN~\cite{zheng2022cmgan}, a state-of-the-art image inpainting model, and finetune {it} from the pretrained CM-GAN weights.}

Specifically, as {shown in} Fig.~\ref{fig:pipeline} (top), our Shadow Analyzer {has} four parts: an Encoder, two parallel cascaded decoders, and a shadow detector. 
The Encoder extracts multi-scale features $F_{e}^i$ ($i \in [1, L]$) and global style code $\boldsymbol{s}$ from $F_{e}^L$. In the parallel decoders, the cascade of global and spatial modulations utilizes the global code $\boldsymbol{s}$ with style code $\boldsymbol{w}$, mapping from noise $\boldsymbol{z}$, to ensure structural coherence and a spatial code for fine-grained detail, producing output features $F_g^{i}$ and $F_s^{i}$. 
For details, including the discriminator architecture, please refer to the Supplementary Materials.

More importantly, we integrate a shadow detector alongside the Spatial Decoder, which processes multi-scale features $F_s^{i}$. This integration, under {the} shadow detection supervision, encourages the encoder and parallel decoders to accurately identify {the} shadow regions. The detector upsamples high-level features (size from 8 to 64) to a uniform size ($64 \times 64$), concatenating them into a single feature map. It comprises a sequence of convolution layers, batch normalization, and GELU layers, interspersed with transpose convolution layers. The final output is a $256 \times 256$ shadow mask, obtained via a sigmoid layer, and subsequently interpolated to match the {size of the} input image.

The Shadow Analyzer is trained with the original combination of adversarial loss, perceptual loss, masked-$R_1$ regularization, and 
{the L1} loss from~\cite{zheng2022cmgan}. Also, we adopt the dice loss~\cite{milletari2016v} to compute the losses between the predicted shadow mask and {the} ground truth.

\subsection{{\stagetwoname}}
\label{sec:synthesizer}

 We illustrate our {\stagetwoname} in the bottom part of Fig.~\ref{fig:pipeline}. It is adapted from an inpainting diffusion model based on the DDPM architecture~\cite{ho2020denoising} trained similarly to the StableDiffusion inpainting model~\cite{Rombach_2022_CVPR}. To support shadow synthesis, it is modified in the following key ways:
(i) We feed an object mask $M_{\widetilde{o}}$
along with the image $I_{o}$ that contains the moved objects as input. This combination enables the model to identify the specific object for which a shadow needs to be synthesized and understand the desired shape of the shadow.
(ii) We incorporate multi-task training into the diffusion model, so that it predicts an additional shadow mask $M_{s}$ at each diffusion step.
(iii) To transfer the shadow knowledge from {\stageonename} and align the dimensions from $F_{ms}$, the multi-scale features with dimensions of $[N, 1348, 32, 32]$, to the original text embeddings, we insert an adaptor $T(\cdot)$ with
a 2D convolution layer followed by a 1D convolution layer. 
Additionally, we employ
a Multilayer Perceptron (MLP) layer to increase 
the embedding dimension from 1344 to 2048, so that
the final shadow embedding $E_s$ has dimensions of $[N, 1024, 2048]$, where $N$ denotes the batch size. 
We then inject it into the diffusion model through 
a
cross-attention mechanism. 

The loss function for training our {\stagetwoname} is
\begin{equation}
\mathcal{L}_{syn}=\mathbb{E}_{T, \epsilon \sim \mathcal{N}(0,1)}\left[\left\|\epsilon-\epsilon_\theta\left(I^t_{o}, M_{\widetilde{o}}, M_{\widetilde{s}}, t, T(F_{ms}) \right)\right\|_2^2\right],
\end{equation}
where $\epsilon \sim \mathcal{N}(0,1)$ is an initial noise, $\epsilon_\theta$ denotes the denoising U-Net, and $I^t_{o}$ is a noisy version of $I_{o}$ at timestep $t$. Note that $M_{\widetilde{s}}$ is an optional shadow mask, which is further explained in the supplemental materials.

\begin{figure}[tp]
    \centering
	\includegraphics[width=0.98\linewidth]{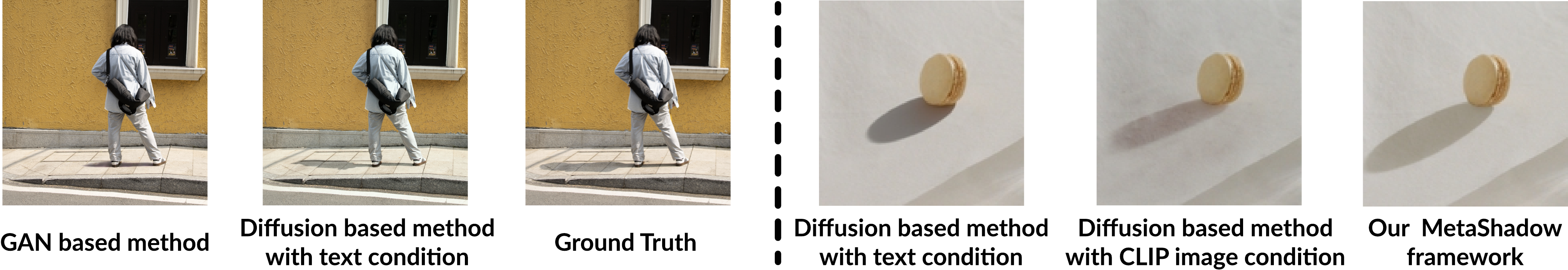}
    \vspace{-2mm}
	\caption{Respective limitations of GAN-based and diffusion-based methods on shadow synthesis. For more discussion, please see Sec.~\ref{sec:ablation}.}    \vspace{-5mm}
	\label{fig:motivations}
\end{figure}

{\bf Discussion.}
In this work, we unveil a unique insight: the integration of GANs and diffusion models overcomes their respective limitations, enabling a more controlled and realistic object-related image editing. As shown in Fig.~\ref{fig:motivations}, we find the GAN excels in effectively and efficiently detecting and removing specific shadows but struggles with synthesizing reasonable shadow shapes~\cite{zheng2022cmgan,hong2022shadow}, as shown in the left part of Fig.~\ref{fig:motivations}, whereas diffusion models excel in generating realistic contents but lack precise control for the light direction, color, and intensity of the shadow as shown in the right part of Fig.~\ref{fig:motivations}. By conditioning diffusion models with GAN features, we can enable controllable and realistic object-centered shadow editing.

\subsection{Training Strategies}

{\bf Multi-source Dataset Training.} As mentioned in Sec.~\ref{sec:datasets}, existing shadow-related datasets are limited at scale. As we aim 
for more general and realistic image editing,
we employ multiple
datasets to train the {\modelname} framework. 

For {\stageonename}, we adopt two types of datasets.
(i) \textbf{Datasets with full annotations}: DESOBA~\cite{hong2022shadow} and our MOS dataset contain shadow images, object masks, shadow masks, and shadow-free images.
    (ii) \textbf{Datasets with partial annotations}: ISTD+~\cite{Le2019Shadow} and SRD~\cite{qu2017deshadownet} contain shadow images, shadow masks, and shadow-free images. When training on this dataset type,
we simply feed an empty object mask and make the model predict general shadows and shadow masks. Also, we randomly make the object mask empty for datasets with full annotations in training.
With this data combination strategy,
{\stageonename} is able to detect an object's shadow 
with a non-empty object mask, and detect general cast shadows with an empty object mask.

For training the {\stagetwoname}, we combine MOS, DESOBA~\cite{hong2022shadow}, and Shadow-AR~\cite{liu2020arshadowgan}. During the training, we randomly choose another object as the reference when there are multiple objects in the image. For the MOS dataset, we also use the moved object as the reference.

{\bf Shadow-Specific Data Augmentations.} We perform three shadow-specific data augmentations to improve the model's generalizability and controllability: (i) Random shadow intensity augmentation,  (ii) Curve-based shadow color grading, and (iii) Random shadow dropping. For more details, please refer to the Supplementary Materials.

\section{Experiments and Results}
\label{sec:results}

{\bf Implementation details.} 
We train {\modelname} in two stages. In Stage I, we train \stageonename~for 100 epochs with a learning rate of 0.001 and batch size of 16. We iterate on the DESOBA dataset~\cite{hong2022shadow} 10 times to balance the number of samples in the multi-dataset training. The training and inference resolution are both $512 \times 512$. In Stage II, we freeze the \stageonename~and fine-tune the diffusion U-Net. We also train the Adaptor in \stagetwoname~from scratch. The inputs and outputs of \stagetwoname~are all at $128 \times 128$ resolution with a batch size of 64. Additionally, we employ different learning rate strategies for the U-Net and the Adaptor. The learning rate for the U-Net begins at $1e-4$ and is multiplied by 0.01 after 200 epochs (400 epochs total), while the learning rate for the Adaptor remains constant at $1e-4$ to strengthen its ability to gain shadow knowledge. We apply random horizontal flips in both stages to enlarge the training samples. All training stages are conducted on an eight A100-GPU Server with the Adam optimizer.

\subsection{Comparison with Existing Methods} \label{sec:exist}
Though no existing methods aim for the same goal as ours,
we evaluate our {\modelname} on four benchmark datasets, including SOBA~\cite{Wang_2020_CVPR}, the DESOBA~\cite{hong2022shadow} test set, Moving DESOBA, and Video DESOBA with different methods for different sub-tasks: object-centered shadow detection, removal, and synthesis.

{\bf Evaluation on object-centered shadow detection.} To evaluate this task, we utilize the common mIoU metric at different sizes, following the COCO~\cite{cocodataset} definitions with an additional extra small category. Tab.~\ref{tab:detection} reports the comparison results on the SOBA test set on shadow segmentation quality. 
As SSIS~\cite{Wang_2021_CVPR,Wang_2022_TPAMI} simultaneously detect all object masks, shadow masks, and their associations, we extract the shadow instance predictions corresponding to each ground-truth object mask for evaluation.
Our {\stageonename} (with or without using MOS Dataset) significantly outperforms both methods across various shadow scales.

\begin{figure*}[htp]
    \centering
	\includegraphics[width=0.86\textwidth]{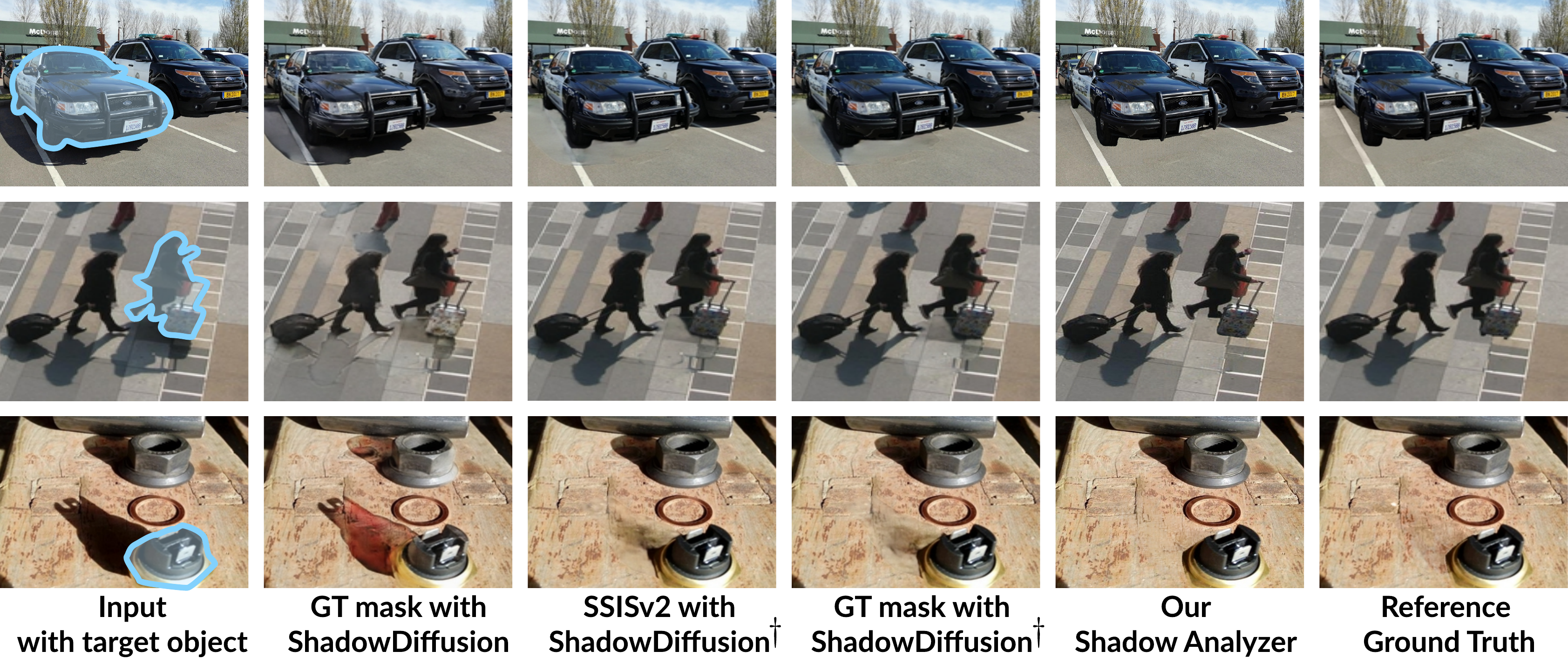}
    \vspace{-1mm}
	\caption{Visual comparison for object-centered shadow detection and removal tasks on the DESOBA test set. $\dagger$ means fine-tuned on our multi-source dataset strategy.  Zoom in to see the details. For more results, please refer to the supplementary materials.}
    \vspace{-2mm}
	\label{fig:sdsr}
\end{figure*}

\begin{figure*}[h]
    \centering
	\includegraphics[width=0.9\linewidth]{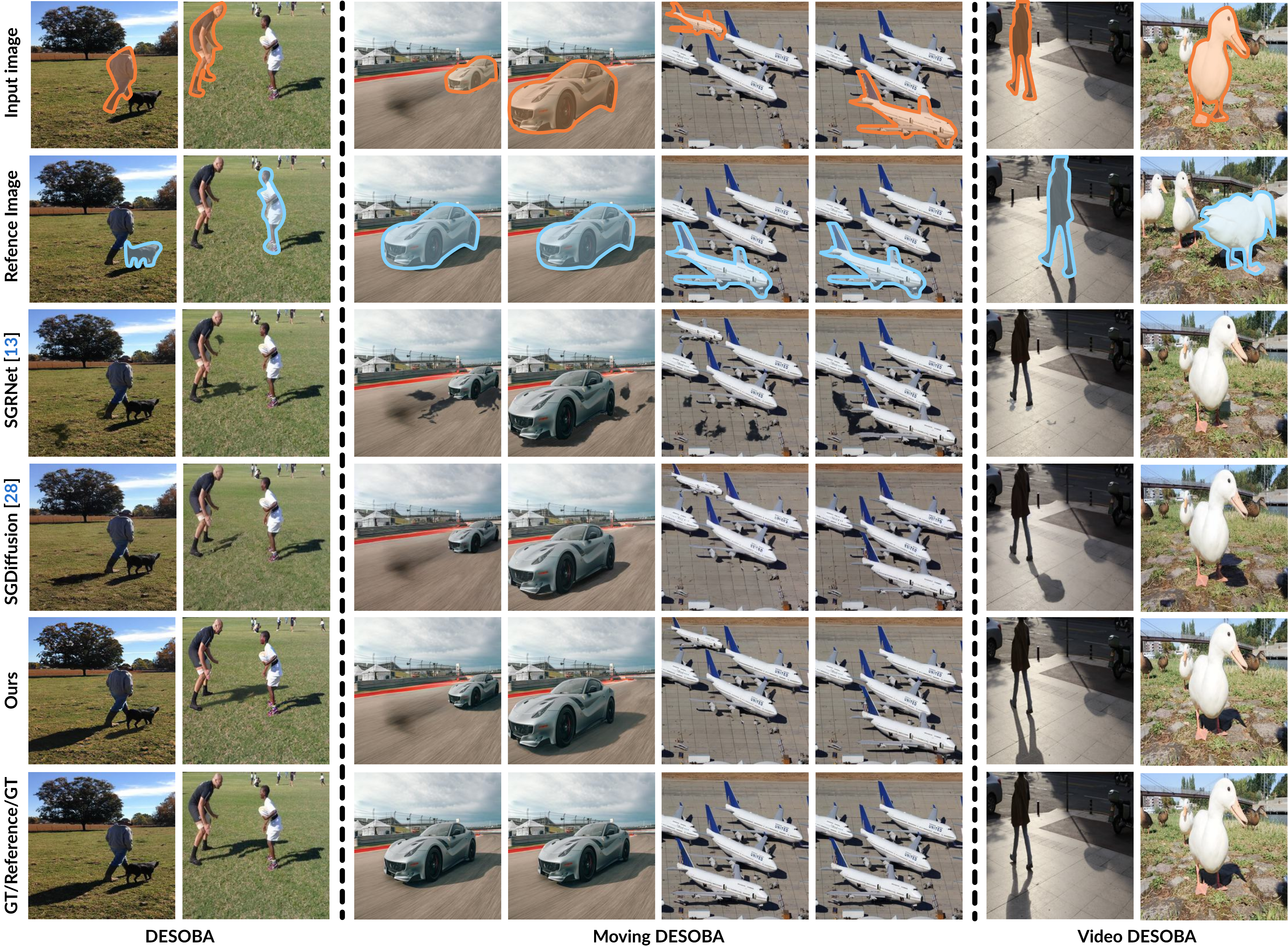}
    \vspace{-1mm}
	\caption{Visual comparison for object-centered shadow synthesis on the DESOBA test set~\cite{hong2022shadow}, our Moving DESOBA test set, and Video DESOBA.  Zoom in to see the details. For more results, please refer to the supplementary materials.}     \vspace{-4mm}
	\label{fig:ss}
\end{figure*}

\begin{figure}[h]
    \centering
	\includegraphics[width=0.8\linewidth]{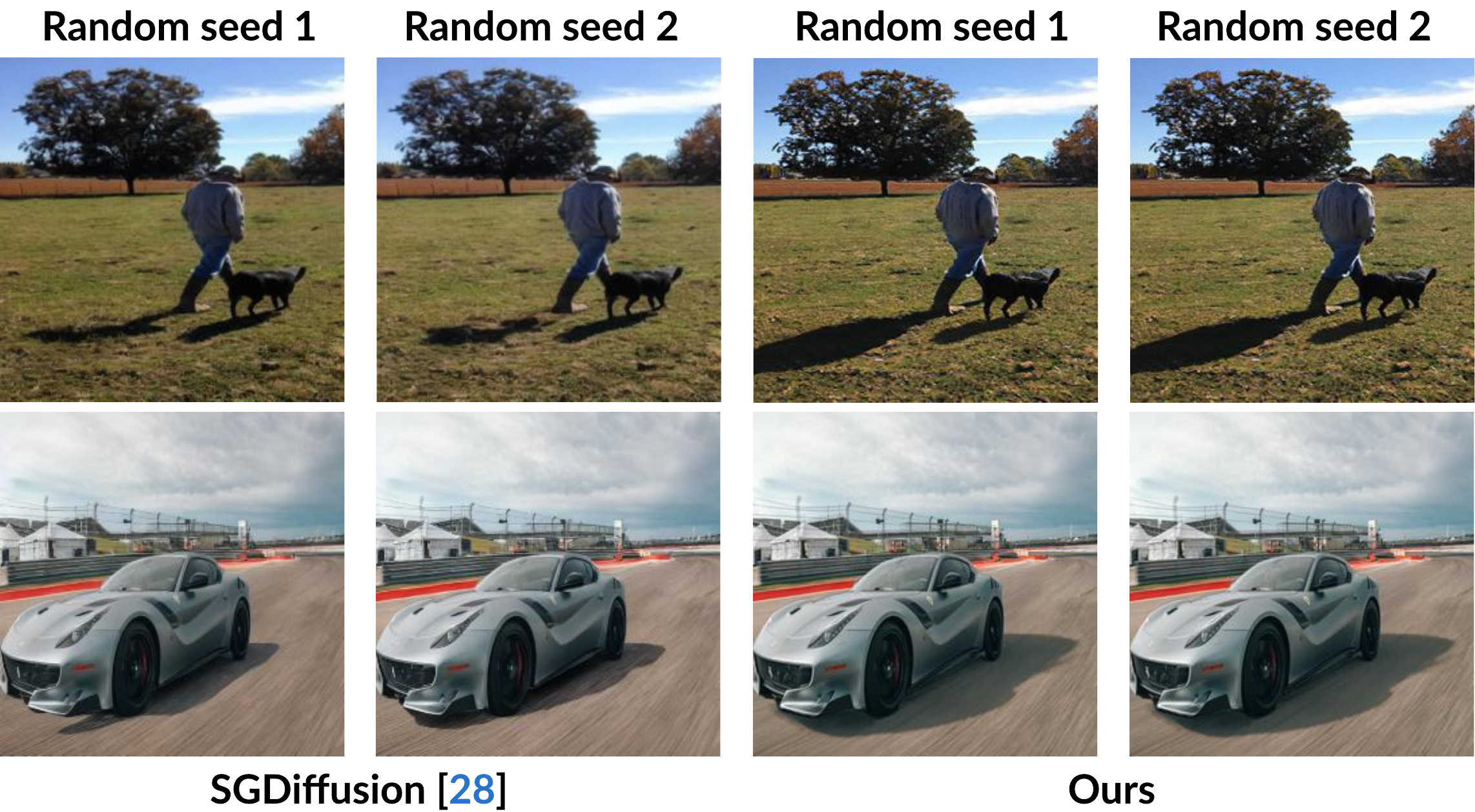}
    \vspace{-4mm}
	\caption{Visual comparison on different random seeds reveals a critical issue with the previous diffusion-based method~\cite{Liu2024SG}: inconsistent shadow generation across various sampled noises. Empirically, our model does not exhibit this weakness.}     \vspace{-4mm}
	\label{fig:consistency}
\end{figure}

{\bf Evaluation on object-centered shadow removal.} We employ Masked MAE, Masked RMSE in LAB color space, Bbox PSNR,  Bbox SSIM, and PSNR to evaluate the performance on this task. For clarity,
\emph{masked} denotes only computing the error inside the ground-truth shadow mask region, and \emph{Bbox} means we compute the error inside the bounding box retrieved from the shadow mask.
We join two recent SOTA methods~\cite{Wang_2022_TPAMI,guo2023shadowdiffusion} in cascade and also finetune the method~\cite{guo2023shadowdiffusion} on our dataset setting for fair comparisons. The results are reported in Tab.~\ref{tab:shadow_removal}. It is evident that training on the original SRD dataset~\cite{qu2017deshadownet} does not result in good generalization on more complex datasets, such as DESOBA~\cite{hong2022shadow}. Furthermore, our method outperforms \cite{guo2023shadowdiffusion} even when using ground-truth masks. We provide comparisons on general shadow removal on the ISTD+~\cite{Le2019Shadow} test set with~\cite{guo2023shadowdiffusion,cun2020towards,liu2023decoupled} in the Supplementary Material. 

Fig.~\ref{fig:sdsr} reveals that before fine-tuning, ShadowDiffusion~\cite{guo2023shadowdiffusion} inadequately recovers shadow regions, leaving residual shadows (1st and 3rd row), or alters other shadows not corresponding to the given object mask (2nd row). After finetuning, it removes shadows but loses detail, causing over-smoothing. 
Additionally, SSISv2's~\cite{Wang_2022_TPAMI} erroneous detections can lead to incorrect shadow region removal (see 3rd row). 
In contrast, our {\stageonename} preserves details under shadows, ensuring high-quality visuals.

\begin{table}[tp]
\renewcommand\tabcolsep{25.0pt}
\resizebox{\columnwidth}{!}{\begin{tabular}{@{}lccccc@{}}
\toprule
Methods & mIoU   & mIoU$_{xs}$ & mIoU$_{s}$ & mIoU$_{m}$ & mIoU$_{l}$ \\ \midrule
SSIS~\cite{Wang_2021_CVPR}  & 51.6 & 37.2      & 46.0     & 66.7     & 81.4     \\
SSISv2~\cite{Wang_2022_TPAMI}  & 55.8 & 42.4      & 49.5     & 70.4     & 82.5     \\
\textbf{Ours} wo MOS  & 67.2 & 54.5      & 70.3     & 79.1     & 86.5    \\
\textbf{Ours}& \textbf{71.0}& \textbf{60.4}& \textbf{72.6}& \textbf{81.1}& \textbf{87.8}     \\ \bottomrule
\end{tabular}}
\caption{Comparison with the SOTA shadow-detection methods on the SOBA test set. Note that SSISv2 automatically detects shadow-object instance pairs in the image, whereas our method uses object masks to detect shadows of objects.}
\vspace{-5mm}
\label{tab:detection}
\end{table}

\begin{table}[t]
\renewcommand\tabcolsep{15.0pt}
\resizebox{\columnwidth}{!}{\begin{tabular}{@{}lcccccc@{}}
\toprule
Method &  \begin{tabular}[c]{@{}l@{}} Masked \\ MAE ↓\end{tabular} & \begin{tabular}[c]{@{}l@{}} Masked \\ RMSE ↓\end{tabular}   & \begin{tabular}[c]{@{}l@{}} Bbox \\ PSNR ↑\end{tabular}  & \begin{tabular}[c]{@{}l@{}} Bbox \\ SSIM ↑\end{tabular}  & PSNR ↑  \\ \midrule
\begin{tabular}[c]{@{}l@{}}ShadowDiffusion~\cite{guo2023shadowdiffusion} \\ with GT shadow mask\end{tabular}     & 60.71 & 17.50 &19.42 & 54.70 & 25.04 \\ \midrule
\begin{tabular}[c]{@{}l@{}}ShadowDiffusion$^\dagger$~\cite{guo2023shadowdiffusion} \\ with SSISv2~\cite{Wang_2022_TPAMI} \\ detected shadow mask\end{tabular} &  39.53  &  12.49  &  23.41     &  68.04  & 39.13   \\ \midrule
\begin{tabular}[c]{@{}l@{}}ShadowDiffusion$^\dagger$~\cite{guo2023shadowdiffusion} \\ with GT shadow mask\end{tabular}     & 35.45 & 11.44 &24.28 & 70.17 & 40.04\\ \midrule
\textbf{Ours}    & \textbf{21.32}  & \textbf{6.62} & \textbf{32.97} & \textbf{96.49} & \textbf{42.20} \\ \bottomrule
\end{tabular}}
\caption{Comparison with SOTA shadow-removal methods on the DESOBA test set. Note that ShadowDiffusion~\cite{guo2023shadowdiffusion} needs a shadow mask as input, so we use the instance shadow mask detected by SSISv2 and the ground-truth shadow mask together to evaluate this method. Note that our {\modelname}, takes an object mask as input.
$\dagger$ denotes fine-tuning on our joint datasets.
} 
\label{tab:shadow_removal}
\vspace{-5mm}
\end{table}
\begin{table}[t]

\footnotesize
\renewcommand\tabcolsep{20.0pt}
\renewcommand{\arraystretch}{1.15}
\resizebox{\columnwidth}{!}{
\begin{tabular}{@{}c|lcccc@{}}
\toprule
 &
  Method &
  \begin{tabular}[c]{@{}l@{}}Global\\ RMSE ↓\end{tabular} &
  \begin{tabular}[c]{@{}l@{}}Local\\ RMSE ↓\end{tabular} &
  \begin{tabular}[c]{@{}l@{}}Bbox\\ PSNR↑\end{tabular} &
  \begin{tabular}[c]{@{}l@{}}Bbox\\ SSIM↑\end{tabular} \\ \midrule
\multirow{4}{*}{ DESOBA} &
  SGRNet~\cite{hong2022shadow}  &
  4.91 &
  56.44 &
  27.29 &
  91.08\\
 &
 SGDiffusion~\cite{Liu2024SG} &
  15.03 &
  64.90 &
  21.53 &
  73.57 \\
 &
 Libcom~\cite{niu2021making} &
  7.88 &
  67.21  &
  23.90 &
  87.48 \\
 &
 \textbf{Ours (256 $\times$ 256)} &
   \textbf{3.12} &
  \textbf{36.84} &
  \textbf{29.16} &
  \textbf{93.56} \\
  &
 \textbf{Ours (128 $\times$ 128)} &
   \textbf{2.93} &
  \textbf{30.92} &
  \textbf{30.73} &
  \textbf{93.49} \\ \midrule
\multirow{4}{*}{\begin{tabular}[l]{@{}c@{}} $\ \ $ Video\\$\ \ $ DESOBA \end{tabular}  } &
  SGRNet~\cite{hong2022shadow} &
  9.89 &
  51.73 &
  20.77 &
  79.14 \\ 
 &
  SGDiffusion~\cite{Liu2024SG} &
  12.40 &
  54.15  &
  36.54 &
  76.12 \\
 &
   Libcom~\cite{niu2021making} &
  12.52 &
  58.29  &
  19.73 &
  75.60 \\
 &
  \textbf{Ours (256 $\times$ 256)} &
  \textbf{8.07} &
  \textbf{36.54} &
  \textbf{23.14} &
  \textbf{82.41} \\
   \bottomrule
\end{tabular}}
\caption{Comparison with the SOTA shadow-synthesis method on the DESOBA subset of the test set with multiple objects in an image and Video DESOBA.  }
\vspace{-5mm}
\label{tab:synthesis}
\end{table}

{\bf Evaluation on object-centered shadow synthesis.} We follow~\cite{hong2022shadow} and utilize Global RMSE, Local RMSE, and our Bbox PSNR and Bbox SSIM as the metrics to evaluate the methods on this task with three baselines~\cite{hong2022shadow,Liu2024SG,niu2021making}. In order to compare on the same image size, we upsample our results to 256x256. Please refer to Supplementary Materials for the upsample process.
As shown in Tab.~\ref{tab:synthesis}, our method demonstrates superior shadow-synthesis quality on the DESOBA and Video DESOBA test sets, significantly reducing the Local RMSE to 36.84. Our {\stagetwoname}~also exhibits enhanced performance for real moving object scenarios in Video DESOBA, reducing the Local RMSE to 36.54. 
Note that since SGRNet~\cite{hong2022shadow} utilizes ground-truth shadow parameters for additional supervision. As our datasets lack these shadow parameters, we cannot fine-tune SGRNet using them. We use the official release of SGDiffusion~\cite{Liu2024SG} and Libcom~\cite{niu2021making}, trained on the 22K DESOBAv2 dataset. However, we found that SGDiffusion tends to darken images and changes the details, leading to lower performance.

To further show the advantage of our method, we provide various visual comparisons in Fig.~\ref{fig:ss},
presenting results on DESOBA test set~\cite{hong2022shadow}, our Moving DESOBA, and Video DESOBA.
As the reference image:
\textit{another} object in the image is used for DESOBA;
the original object before relocation is used for Moving DESOBA; 
the first frame in the video clip is used for Video DESOBA, 
 showing the versatility of our framework.
We compare with SGRNet~\cite{hong2022shadow} using identical reference objects, and SGDiffusion~\cite{Liu2024SG} using the reference shadow masks. 
Even so, our {\modelname} excels in creating realistic shadows for complex shapes, such as airplanes, with precise color (like the first case in Moving DESOBA), intensity, and direction matching, highlighting its effectiveness.
Also, as shown in Fig.~\ref{fig:consistency}, SGDiffusion~\cite{Liu2024SG} generates inconsistent shadows depending on the random seed. On the contrary, our {\modelname} framework achieves consistent shadow synthesis owing to our shadow knowledge transfer mechanism.
See more results, including GIFs of Video DESOBA, in the Supplementary Materials.

\subsection{Analysis on Shadow Knowledge Transfer}~\label{sec:ablation}

Recently, text-to-image generation models~\cite{kawar2023imagic,Rombach_2022_CVPR,li2023gligen} have achieved great success in synthesizing realistic images by injecting the text embedding from the text encoder of a large language model (LLM) (like T5~\cite{raffel2020exploring}) or a vision-language model (VLM) (like CLIP~\cite{CLIP}) into the diffusion model. Some recent methods~\cite{paintbyexample,objectstitch, chen2023anydoor} replace the text embedding with an image embedding or even combine text and image embeddings~\cite{balaji2022eDiff-I}.
Yet, it is hard for LLMs and VLMs to represent/extract fine-grained features for degradation tasks, e.g., shadowed image, as they are typically trained on diverse web-scale data without specific captions for degradation scenarios~\cite{luo2023controlling}.

\begin{table}[tp]
\centering
\renewcommand\tabcolsep{20.0pt}
\resizebox{0.8\columnwidth}{!}{\begin{tabular}{@{}lccc@{}}
\toprule
Method &
  \begin{tabular}[c]{@{}l@{}}Global \\ RMSE ↓\end{tabular} &
    \begin{tabular}[c]{@{}l@{}}Bbox\\ PSNR↑\end{tabular} &
  \begin{tabular}[c]{@{}l@{}}Bbox\\ SSIM↑\end{tabular} \\ \midrule
                     Baseline 1: SSDM-Text~\cite{ho2020denoising,raffel2020exploring} &
  3.36 &
    29.80 &
  92.21 \\
 Baseline 2: SSDM-CLIP~\cite{ho2020denoising,CLIP}  &
  4.51 &
    29.72 &
  93.17 \\
\textbf{Ours} &
 \textbf{2.93} &
  \textbf{30.73} &
  \textbf{93.49} \\  \bottomrule
\end{tabular}}
\vspace{-2mm}
\caption{Ablation study on the DESOBA test set.}
\label{tab:ablation}
\end{table}

\begin{figure}[tp]
\centering

\includegraphics[width=0.8\linewidth]{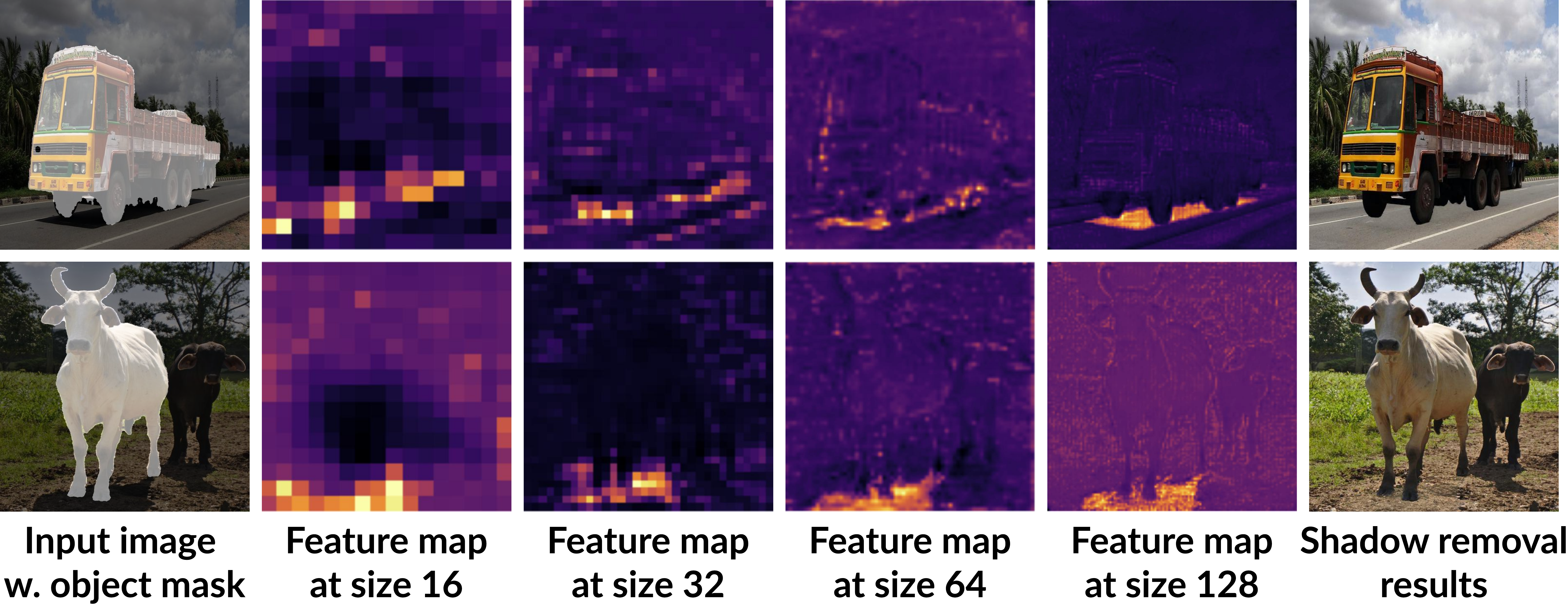}
\vspace{-2mm}
\captionof{figure}{Visualization of the intermediate multi-scale feature maps of our {\stageonename}. Given an object mask, {\stageonename} detects and removes the shadow for specific objects. Thus, the feature maps 
effectively capture
the shadow knowledge, especially in the target shadow regions. }
\label{fig:vis}

\end{figure}

Especially for our {\stagetwoname}, the condition embeddings should ideally include shadow characteristics such as intensity, softness, color, and direction of the original shadow. Thus, a task-specific encoder for shadow feature extraction would better serve the purpose than a general image encoder.
We empirically verify this by comparing the following:
(i) ``SSDM-Text'' representing the Shadow Synthesis Diffusion Model, which has the base architecture~\cite{ho2020denoising} of our {\stagetwoname}~but takes the text embedding from T5~\cite{raffel2020exploring} as the condition with the word ``shadow'' as the text prompt, and
(ii) ``SSDM-CLIP'' representing SSDM with CLIP~\cite{CLIP} image embeddings as condition,
replacing $F_{ms}$. Note that as the original CLIP~\cite{CLIP} image encoder faces the challenge of extracting sufficient shadow knowledge directly from the image, we finetune it with the diffusion model to strengthen its ability. Note that both baselines are trained with the same dataset as Ours until convergence.
Table~\ref{tab:ablation} reports the comparison results, showing that image embeddings are more effective than text embeddings.
Even though CLIP~\cite{CLIP}  is widely used to encode image information~\cite{paintbyexample,objectstitch}, it performs sub-optimally compared to our task-specific shadow knowledge transfer. 

We further visualize feature maps from {\stageonename}, which
distinctly highlight the response of the shadow regions (Fig.~\ref{fig:vis}),
demonstrating the Analyzer's effectiveness in capturing shadow characteristics. 
However, we further observed that larger resolution features gradually include texture information within the shadow region, which is not desired, as we solely want to transfer the shadow properties, not the texture from previous locations. 
Based on this observation, we use
features of varying sizes (16 to 128)  and resize to a uniform $32\times 32$ size as mentioned in Sec.~\ref{sec:synthesizer}. 

Our design additionally offers a significant advantage by requiring only four steps to generate the final result, in contrast to  SSDM's 30 steps or SGDiffusion's~\cite{Liu2024SG} 50 steps. All results presented in this paper and Supplemental Materials are based on this four-step setting.

In the Supplemental Material, we delve deeper into detailed analyses of the dataset and data augmentation techniques through ablation studies, along with more comparisons and visualizations on different sub-tasks as well as the limitation and potential solution.

\section{Conclusion}
\label{sec:conclusion}

In this work, we introduced \modelname, a novel framework for enhancing realism in image editing through advanced shadow manipulation. By integrating \stageonename~for precise shadow detection and removal, and \stagetwoname~for controllable shadow generation, \modelname~achieves a significant leap in object-centered image processing. This synergy ensures that shadows are not only realistic but also contextually harmonized with the scene, eliminating the need for complex systems requiring lighting and geometry parameters. 
Our evaluations demonstrate \modelname's superior performance over existing methods, with notable improvements in object-centered shadow detection, removal, and synthesis. This framework enables various image-editing tasks, such as object removal, relocation, and insertion, showcasing its potential to advance object-centered image editing techniques.

{
    \small
    \bibliographystyle{ieeenat_fullname}
    \bibliography{main}
}

\end{document}